\documentclass{article}
\usepackage{times}
\usepackage{epsfig}
\usepackage{graphicx}
\usepackage{amsmath}
\usepackage{amssymb}
\usepackage{bm}
\usepackage{multicol}
\usepackage{authblk}
\usepackage[utf8]{inputenc}
\usepackage{indentfirst}
 \usepackage{float}
\usepackage[super]{nth}
\usepackage[nottoc]{tocbibind}
\usepackage{caption}
\usepackage{dblfloatfix}
\usepackage{soul, xcolor}
\usepackage[left=1.8cm, right=1.5cm,top = 2cm,bottom=3cm]{geometry}
\usepackage{abstract}
\usepackage{hyperref}
\hypersetup{pagebackref=true,breaklinks=true,colorlinks,bookmarks=false}

\DeclareUnicodeCharacter{0301}{*************************************}

\begin{document}

\title{\Large \bf A Spacecraft Dataset for Detection, Segmentation and Parts Recognition}
\author[1]{Hoang Anh Dung}
\author[2]{Bo Chen}
\author[2]{Tat-Jun Chin}
\affil[1]{The University of Adelaide}
\maketitle

\begin{multicols}{2}
\paragraph{}
\centerline{\large\bf Abstract}%
{\it
Virtually all aspects of modern life depend on space technology. Thanks to the great advancement of computer vision in general and deep learning-based techniques in particular, over the decades, the world witnessed the growing use of deep learning in solving problems for space applications, such as self-driving robot, tracers, insect-like robot on cosmos and health monitoring of spacecraft. These are just some prominent examples that has advanced space industry with the help of deep learning. However, the success of deep learning models requires a lot of training data in order to have decent performance, while on the other hand, there are very limited amount of publicly available space datasets for the training of deep learning models. Currently, there is no public datasets for space-based object detection or instance segmentation, partly because manually annotating object segmentation masks is very time consuming as they require pixel-level labelling, not to mention the  challenge of obtaining images from space. In this paper, we aim to fill this gap by releasing a dataset for spacecraft detection, instance segmentation and part recognition. The main contribution of this work is the development of the dataset using images of space stations and satellites, with rich annotations including bounding boxes of spacecrafts and masks to the level of object parts, which are obtained with a mixture of automatic processes and manual efforts. We also provide evaluations with state-of-the-art methods in object detection and instance segmentation as a benchmark for the dataset. The link for downloading the proposed dataset can be found on \href{https://github.com/Yurushia1998/SatelliteDataset}{https://github.com/Yurushia1998/SatelliteDataset}.
}

\section{Introduction}
Space technologies play a vital role in many critical applications today: communication \cite{Satellite_communication}, navigation \cite{Satellite_navigation} and meteorology \cite{satellite_meteorology} are some prominent examples, thanks to the development of computer vision and machine learning techniques. Within the last two decades, there has been a wide range of machine learning-based applications deployed in the space industry, such as self-navigation system for collision avoidance \cite{Uriot2020SpacecraftCA}, health monitoring of spacecrafts~\cite{health_monitering}, and asteroid classifications \cite{Asteroid_classification}, just to name a few. Accompanying the development of space technologies is an increase in demand of space datasets, as most of state-of-art models for space technologies are using deep learning-based methods, which require a significant amount of annotated data for supervised training in order to have good performance. However, one challenge that hinders the advancement of these space technologies is the lack of publicly available datasets, due to sensitivity in the space area and the high cost of obtaining space-borne images.


One important technology in many space applications is the accurate localisation of space objects via visual sensor, such as object detection and segmentation in images, because localisation is a key step towards vision-based pose estimation which is critical for tasks like docking~\cite{docking}, servicing~\cite{reed2016restore}, or debris removal~\cite{forshaw2016removedebris}. However, a severe challenge for space-based object detection and instance segmentation is the lack of accessible large datasets that have been well annotated. There has been some large scale segmentation dataset such as COCO\cite{coco_dataset}, ImageNet \cite{5206848}, Pascal VOC \cite{Pascal_VOC} including masks of a large number of classes for daily life objects and human parts, but there is not any specialized datasets segmenting space objects such as satellites, space station, spacecrafts or other Resident Space Objects (RSOs). The closest and the largest datasets related to this topic so far are the Spacecraft PosE Estimation Dataset (SPEED)~\cite{SPEED_dataset} and the URSO dataset~\cite{URSO_dataset}. However, these datasets are focused on pose estimation and do not provide any segmentation annotations. 

Since pixel-level mask are required as ground truth for training, building a segmentation dataset for any new domain, can be very time-consuming. For example, powerful interactive tools~\cite{test_tools} are adopted for annotating the MS COCO dataset \cite{coco_dataset}, but it still takes minutes for an experienced annotator labeling one image \cite{coco_dataset}. As the large amount of parameters in modern neural networks often require being trained on fairly large datasets in the scales from thousands to millions, the total amount of effort it takes to develop such a dataset remains dauntingly expensive. 



To reduce the cost and manpower required for image mask annotation, there has been many researches trying to automate or semi-automate the annotation process using unsupervised approaches, such as interactive segmentation~\cite{scribble_refiner_scheme,interactive} where human annotators use a model to create sample masks and interact with the samples iteratively to refine it, or weakly supervised annotation methods~\cite{scribble_2,extreme_points} where users only need to provide a ‘weak’ annotation giving minimum information about the mask of the images. Another line of works trying to circumvent the expensive cost of annotation is to rid the need of annotation at all via self-supervised learning, such as \cite{self_transform, self_weather}. However, self-supervised learning based methods tend to be inferior in detection and segmentation tasks compared to their supervised counterparts.




\paragraph{Our contributions}

In this work, we aim to contribute to space-based vision researches by creating a new public available space images dataset, as the first step in a long term goal to develop new machine learning algorithms for spacecraft object detection, segmentation and part recognition tasks.

As spaces images are often considered as sensitive data, there are not many real satellite images are publicly available. To enrich our image dataset, we collect 3117 images of satellites and space stations from synthetic or real images and videos. We then use a bootstrap strategy in the annotation process to maximally reduce manual efforts required. We first adopt an interactive method for manual labelling at a small scale, then utilised the labelled data to train a segmentation model for automatically producing coarse labels for more images, which have subsequently gone through manual refinement via the interactive tool. As more finely annotated images are produced, this process repeats and scales up until we finally produces the whole dataset. 

To provide a benchmark for our dataset we conduct experiments using state-of-the-art detection and segmentation methods. The performance of our dataset in comparison to popular datasets such as Cityscapces~\cite{Cordts2016Cityscapes} and Pascal VOC~\cite{Pascal_VOC} indicates that space-based semantic segmentation is a challenging task for models designed based on on-Earth scenarios and poses a open domain for future research.

\section{Related works}
Image segmentation is a topic that has been studied for a number of decades in the field of computer vision, which has recently regained significant attention due to the success of deep learning. Consequently, the demand for annotated datasets has grown rapidly. There has been various researches that focus on minimizing the cost of training segmentation models via either improving the data annotation techniques or reducing the reliance on labelled data in training. We briefly review notable techniques in data annotation and self-supervised learning.

\paragraph{Data annotation} 
To minimise the amount of human input in image mask annotation, various techniques have been proposed. Maninis~\cite{extreme_points} use extreme points of the objects that are to be segmented (points to the top, the bottom, the left-most and the right-most on the boundary of the object) as a annotation signal. Each extreme point is converted to a 2D Gaussian heatmap and concatenated to the input image as an extra channel of features. The model then learns to utilise this information to produce an accurate mask. Scribble-based techniques~\cite{scribble_2, scribble_refiner_scheme} on the other hand, use a scribbled-based input as an annotation signal. Unlike extreme points or bounding boxes, scribbles does not give information about the object location, instead it provides information about color and intensity of the objects to be segmented. The model then propagates this object specific information from scribbles to other pixels and estimates the object masks. Other notable techniques in mask annotation include bounding box input~\cite{bounding_box}, object centre input~\cite{center}, polygon input~\cite{polygon, castrejon2017annotating} and interactive approaches~\cite{polygon_tool, interactive}.

\begin{figure*}[t] 
    \centering
    \includegraphics[width=0.9\linewidth]{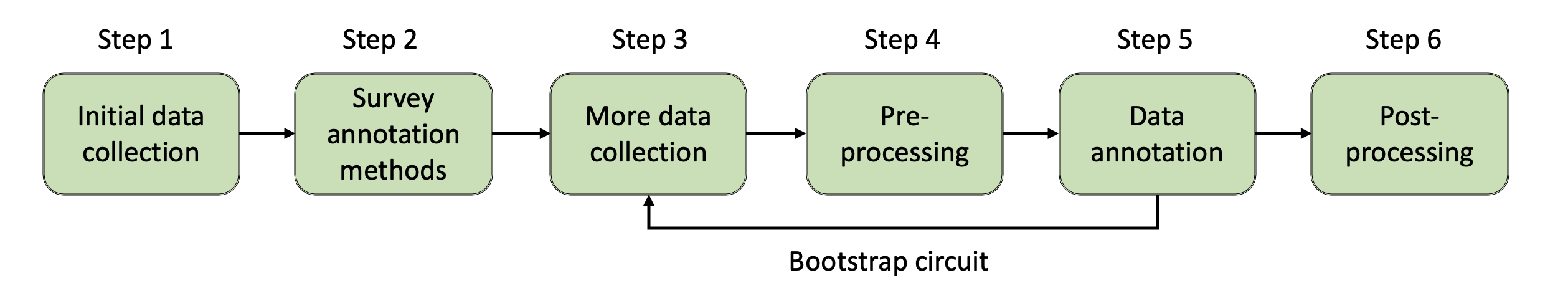}
    \caption{Process from data collection to image segmentation. }
  \label{fig:pipeline}
\end{figure*}

\paragraph{Self-supervised learning} This is a topic that has recently gained popular attention as it addresses the lack of training data problem in deep learning methods. There has also been efforts for image segmentation based on self-supervised learning. The main idea of this approach is 
to facilitate the model to learn information about inherent structures within images by training the model with the same input data with different representation or augmentation. In 2019, Larsson ~\cite{self_weather} used a k-mean clustering to get predicted labels for pixels of 2 different images from the same scene with different weather condition. 
It uses a correspondent loss of the differences in segmentation of both images to learn meaningful features, as they should have the same labels. Another work~\cite{self_transform} uses a self-supervised equivariant attention mechanism to 
provide additional supervision signal in semantic segmentation. In this work, instead of using images of same scenes at different weather condition, it applies affine transformations on input images and uses an equivariant cross regularization loss to encourage feature consistency in learning. 

\section{Building the dataset}

In this section we describe our methodology for data collection and mask generation throughout the dataset development process.
We collect a large number of synthetic and real images from the internet. We then use a bootstrap strategy to effectively reduce the amount of manual labor required for data annotation. We further perform a post-process step to remove similar or identical images, remove texts or refine low-quality annotations. 
Figure~\ref{fig:pipeline} illustrates the step-by-step process of our methodology.


\subsection{Step 1: Initial data collection} For the dataset to be useful for training practical models, we need to collect a significant amount of data. However at this step only a small amount was needed since we would only use them to test viable annotation methods that are easy to operate and are able to produce satisfactory masks. 

\subsection{Step 2: Surveying annotation methods}
Using the initial amount of data from previous step, we conduct experiments on current various state-of-the-art models and tools for image segmentation. Decomposing spacecrafts into parts requires specialised domain knowledge. For practical reasons, we opt to segment spacecrafts into 3 parts that are commonly observable and easily identified, namely solar panel, main body and antenna.

\begin{figure}[H] 
    \centering
    \includegraphics[width=\linewidth]{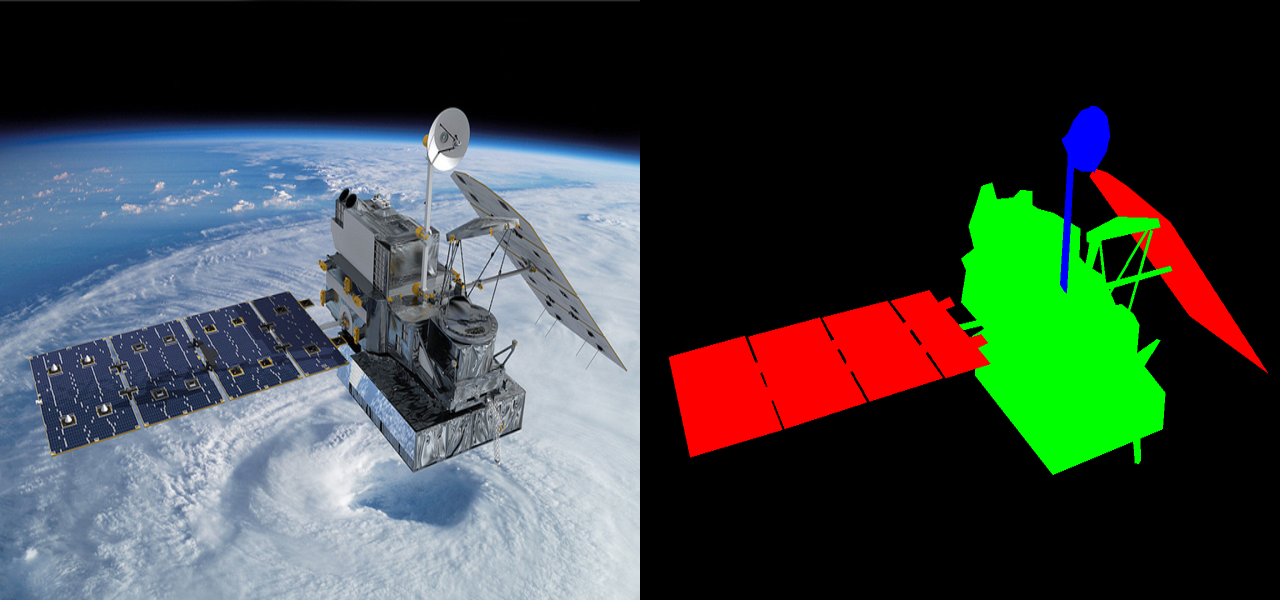}
    \caption{Example of an image collected and its annotated masks. Red mask: solar panel; blue mask: antenna; green mask: main body. }
    \label{fig:mask_example}
\end{figure}

Among the surveyed methods, self-supervised \cite{self_transform,self_weather} and weakly supervised segmentation methods  \cite{scribble_2} have advantages of low human interaction and labor requirements per images. However, their performance are no where near satisfactory, as the satellites often have a lot of unorthodox and small parts. Another drawback of self-supervised or weakly supervised approaches is that it is highly inefficient to further refine their output predictions. On the other hand, interactive segmentation methods such as \cite{extreme_points} have advantages of allowing users to improve the mask bit by bit with manual inputs, which are much more suitable for the purpose of this work. 
After testing various methods, we decided to use Polygon-RNN++~\cite{polygon_tool}, an improved version of Polygon-RNN~\cite{castrejon2017annotating}. 
This model allows us to break down the object into small convex areas. We can then label each of these convex area with polygons manually based on their position on the spacecraft, and the mask will then be created. The model also allows users to freely modify the mask at pixel level by adding or removing key points. Figure~\ref{fig:mask_example} is an example image with masks of each part, labelled using Polygon-RNN++. 

\subsection{The bootstrap circuit}

From step 3 to 5, we employ a bootstrap strategy to make the labelling process semi-automatic, via utilising already annotated images to train a segmentation model to generate initial mask predictions. We first collect more images to expand our data base. A pre-processing step was then conducted to remove similar or duplicate images. Lastly we train segmentation models to predict the initial masks and refine them using Polygon-RNN++ as described in step 2. This bootstrap circuit is repeated and as the training data grows, the segmentation models also improve, which in turn further lowers the cost of labelling more data. 

\paragraph{Pre-processing}
In order to remove duplicate or near-duplicate images due to difference in size, resolution and augmentation, we used Agglomerative Clustering~\cite{day1984efficient} implemented in sklearn~\cite{sklearn_api}, combined with a simple searching algorithm.
For each image $I_i$, we create a feature vector $\bm f^{(i)} = [f_1^{(i)},...,f_M^{(i)}]^T \in \mathbb R^M$ using the color histogram of the image. We then rrun the clustering algorithm based on the chi-square distance
\begin{equation}
    d(I_i, I_j) = \frac{1}{2}\sum_{k=1}^M{\frac{(f_k^{(i)}-f_k^{(j)})^2}{f_k^{(i)}+f_k^{(j)}}} 
\end{equation}
between two images $I_i$ and $I_j$.


After the clustering algorithm groups similar images into clusters, 
we use a searching algorithm to find the top $n$ couples of images with highest similarity based on the chi-square distance, and manually remove those that are nearly the same. 


\begin{figure*}[t]
 \centering
  \includegraphics[width=0.9\linewidth]{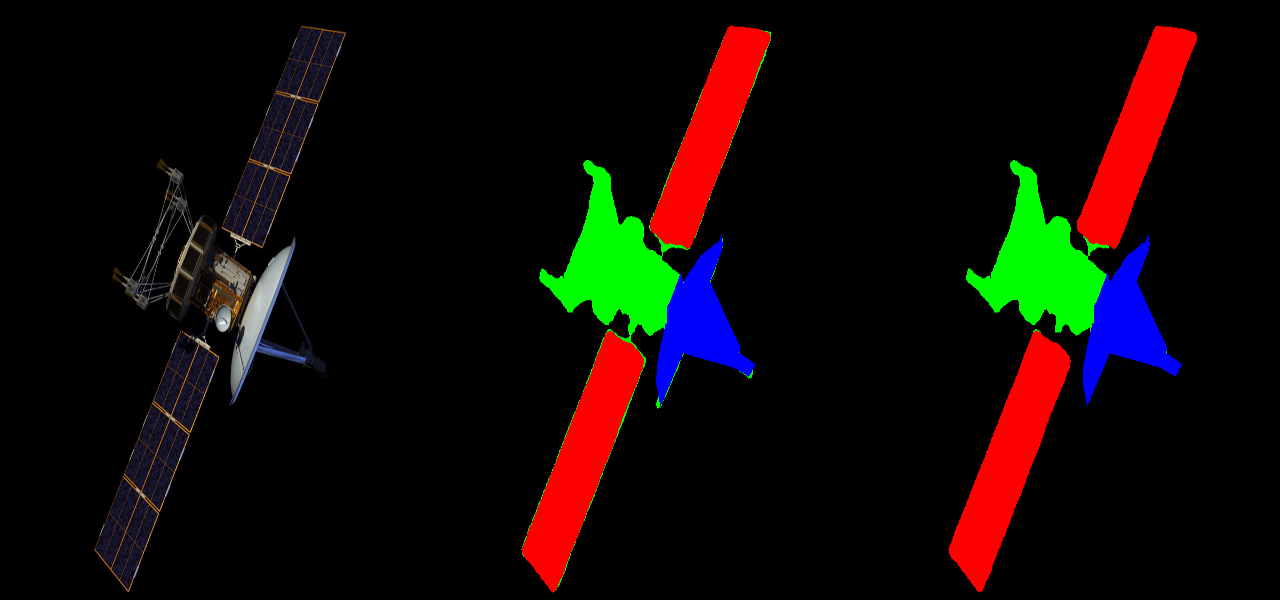}
  \caption{Comparing masks before and after manual refinement. Left: input image. Centre: assembled masks from model predictions. Right: masks after manual refinement. The models here are trained on 1003 annotated images. }
  \label{fig:pred_vs_refine}
  \captionsetup{justification=centering}
\end{figure*}

\paragraph{Data Annotation}


We label the initial batch of images manually as described in step 2. For all further iterations, we leverage existing annotations to support the labelling process by training state-of-the-art models for producing initial mask predictions for different parts. 

We use the DeepLabV3~\cite{DeepLabV3} architecture and refined initial weights pretrained on ImageNet~\cite{5206848} on our dataset. We train 3 different models to predict: full mask of spacecrafts, mask of the solar panels, and mask of antennae. For each image, we then adopt good predictions of the parts, assembled them, and manually refine the final mask. Figure~\ref{fig:pred_vs_refine} compares the predicted masks and the refined masks of an example image. 

As we accumulate more annotated images, the trained models were further refined with the latest dataset to improve future predictions, which in turn, reduces manual efforts and speeds up the labelling of new images. 


\subsection{Step 6: Post-processing}
After we got a sufficient number of masks from the bootstrap circuit, we start going back to problematic images that are marked as requiring further processing. There are images that has text need to be removed, similar images that failed to be filtered in step 4, images that are deemed to be too difficult to identify even with human vision, etc. We also go back to re-mask those that we deem low in quality. Once we obtain the mask labels, we compute tight bounding boxes of the spacecrafts for each image.

\section{Dataset statistic}

\begin{figure}[H]
 \centering
  \includegraphics[width=\linewidth]{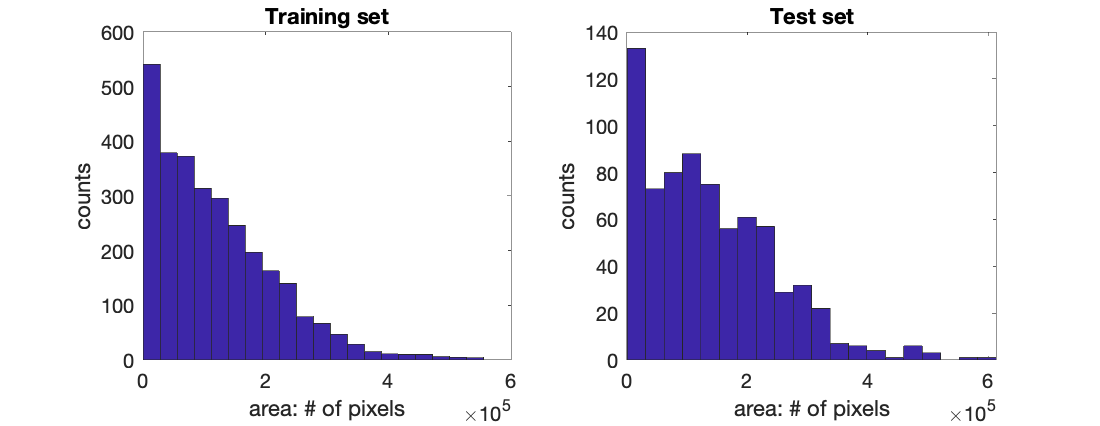}
  \caption{Histograms of the spacecraft mask areas in the training and test set.}
  \label{fig:hists}
  \captionsetup{justification=centering}
\end{figure}

The final dataset consists of 3117 images with uniform resolutions of $1280\times720$. It includes masks of 10350 parts of 3667 spacecrafts. The spacecraft objects are also various in range, they can be as small as 100 pixels to as large as occupying nearly the whole images. On average, each spacecraft takes up an area of 122318.68 pixels, while each part of antenna, solar panel and main body occupies areas of 22853.64, 75070.76 and 75090.92 pixels, respectively. For the purpose of standardising benchmarking segmentation methods, we divide our dataset into a training and a test subsets, which consists of 2516 and 600 images respectively.  Figure~\ref{fig:hists} provides the distribution of spacecraft sizes in the training and test sets. 



\section{Experiments}
In this section we conduct various experiments to benchmark our dataset with state-of-the-art models in tasks including object detection, instance segmentation and semantic segmentation. 

\subsection{Spacecraft object detection}

To serve as the benchmark for spacecraft object detection, we train object detection models such as various versions of YOLO~\cite{yolov3,yolov4} or EfficientDet~\cite{EfficientDet}. We use the same training and evaluation settings as in their original codes, except that we change the training input size of YOLO models to 1280x1280, while testing size remains the same (640x640). We initialise YOLO models with ImageNet pretrained weights, and COCO pretrained weights for EfficientDet models. All models were trained for 20 epochs (around 50000 iterations for EfficientDet). For evaluation metrics, we use meanAP~\cite{coco_dataset} and meanAP50~\cite{coco_dataset}. As shown in Table~\ref{tab:detection}, our experiments suggest that EfficientDet has better detection performance than the YOLO variants in the space-based scenario.
\begin{table}[H]
 \centering
 \begin{tabular}{l| c c} 
 \hline
    Model &  mAP & AP50 \\
 \hline
  YoloV3   & 0.700 & 0.852 \\ 
 YoloV3-spp &  0.736 & 0.868  \\ 
 YoloV4-pacsp  &  0.807 & 0.896  \\ 
  YoloV4-pacsp-x &  0.788 & 0.896  \\ 
  EfficientDet 7D  &  \textbf{0.880} & \textbf{0.904}\\ 
 EfficientDet 7DX  & 0.707 & 0.902\\ 
 \hline
\end{tabular}
\caption{Mean AP50:95 and AP50 of different models on datasets }
 \label{tab:detection}
\end{table}

\subsection{Spacecraft instance segmentation}

To benchmark our dataset with state-of-the-art models for spacecraft instance segmentation, 
We use a variety of segmentation models to test their performances on our dataset, including HRNet~\cite{HRNet}, OCRNet~\cite{OCR}, OCNet~\cite{OCNet}, ResneSt~\cite{Zhang2020ResNeStSN} and DeepLabV3+ Xception~\cite{DeepLabXception}. 
We maintain as much as possible the original training settings as in the respective papers, with some small dataset or hardware specific adaptations. All models are trained with around 40 epochs (For HRNet, AspOCNet and OCRNet, we train them with 13000-15000 iterations). The training batch sizes for DeepLabV3+ Xception were 8 while the rest of models had batch size 4. We use pixel accuracy (PixAcc)~\cite{Pixel_accu_definition} and mean intersection-of-union (mIoU)~\cite{IoU_definition} to compare model performances.


For ResneSt models, we use 3 different backbone with DeepLabV3, including ResneSt101, ResneSt200 and ResneSt269, with DeepLabV3 and an extra auxillary header as segmentation head, so the loss is a weighted combination of losses between DeepLabV3 header output and auxillary header output. All input training images are cropped to size $480\times480$, except for ResneSt269 which had input size $420\times420$.

For DeepLabV3+ with Xception, we use Resnet101 as backbone for the model and crop the input image to size $513\times513$. Similar to ResneSt models, we use full images instead of a crop for testing. 

For HRNet, we use model HRNet48W OCR with pretrained HRNet48W as backbone. Similar to ResneSt models, we also use an extra auxillary head and weighted auxillary loss in this model. All original images and masks are resized to $1024\times512$ for training and $2048\times1024$ for validation, similar to how HRNet processes Cityscape Images. On the other hand, OCNet and OCRNet use Imagenet pretrained weights Resnet101 as backbone. All models use SGD as optimizer with weight decay. 


Table~\ref{tab:satellite_segmentation} reports the segmentation results across different methods on our dataset. Because a significant part of the image is background, which does not contribute much meaningful information and can affect the result of model evaluations, we represent two mIoU results of including and excluding the background class. 

\begin{table}[H]
 \centering
 \begin{tabular}{l| c c c} 
 \hline
    Model & PicAcc & mIoU & \vtop{\hbox{\strut mIoU}\hbox{\strut (No BG)}}\\
 \hline
 DeepLabV3+Xception & 0.965 & 0.78 & 0.714\\ 
 ASPOCNET &  0.972 & 0.803 & 0.744 \\ 
 OCRNet &  0.972 & 0.802 & 0.742 \\ 
 HRNetV2+OCR+ & 0.974 & 0.797 & 0.735 \\ 
 ResneSt101  & 0.977 & 0.822 & 0.767  \\ 
 ResneSt200  & \textbf{0.978} & \textbf{0.838} & \textbf{0.79} \\ 
 ResneSt269  & 0.977 & 0.835  & 0.786 \\ 
 \hline
\end{tabular}
\caption{Performances of different state-of-the-art models for whole spacecraft segmentation.}
 \label{tab:satellite_segmentation}
\end{table}

\begin{table}[H]
 \centering
 \begin{tabular}{l| c c c} 
 \hline
    Model & Body & Solar panel & Antena \\
 \hline
  DeepLabV3+ xception  & 0.767 & 0.802 &  0.575 \\ 
 ASPOCNET &  0.800 & 0.842 & 0.588 \\ 
 HRNetV2+ OCR+ &  0.814 & 0.856 & 0.533 \\ 
  OCRNet &  0.803 & 0.839 &  0.585 \\ 
  ResneSt101  &  0.834 & 0.868 & 0.600\\ 
 ResneSt200  & \textbf{0.842} & \textbf{0.878} & 0.640\\ 
 ResneSt269  &  0.830 & 0.870 & \textbf{0.65} \\ 
 \hline
\end{tabular}
\caption{mIoU by spacecraft parts in different models }
 \label{tab:parts_segmentation}
\end{table}

\begin{table*}[t]
 \centering
 \begin{tabular}{l| c c c c} 
 \hline
    Model & Pascal VOC  & City. &  City. val & Ours \\
 \hline
  DeepLabV3+ Xception & 0.890 & 0.821 & 0.827 & 0.714  \\ 
 ASPOCNET & -    & 0.817 & -    & \textbf{0.744}  \\
 OCRNet & 0.843    & 0.824 & 0.806 & 0.742    \\ 
 HRNetV2+OCR+  & 0.845 &  \textbf{0.845} & 0.811 & 0.735 \\ 
 ResneSt101  & -   & 0.804  & -  & 0.767     \\
 ResneSt200  & -  & 0.833  & 0.827 & 0.790   \\ 
 \hline
 Average & 0.859 & 0.824 & 0.818 & 0.749 \\
 \hline
\end{tabular}
\caption{mIoU of state-of-the-art models across different datasets, treating spacecraft parts as classes of objects.}
 \label{tab:parts_as_classes}
\end{table*}

\subsection{Spacecraft parts segmentation}

\begin{figure*}[t]
 \centering
  \includegraphics[width=\textwidth]{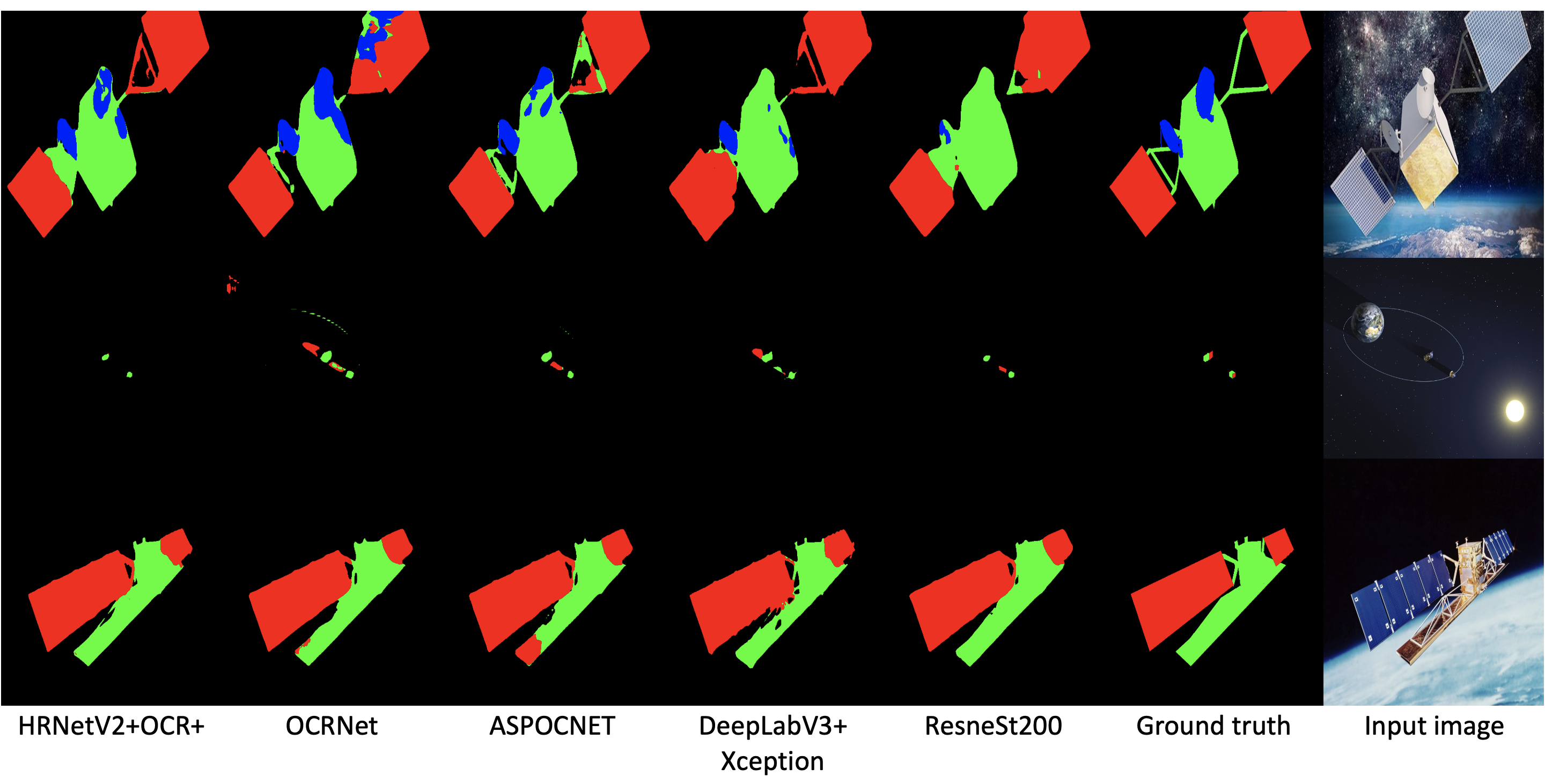}
  \caption{Predicted parts masks from different models.}
  \captionsetup{justification=centering}
  \label{fig:qua_results}
\end{figure*}

Table~\ref{tab:parts_segmentation} shows the performance of state-of-the-art models segmenting spacecraft parts on our dataset. The body and the solar panel of the spacecraft have been decently segmented as reflected by mIoU. The performance for antenna on the other hand, is fairly poor since they are quite unorthodox in shape and difficult to identify. Also, it is noticeable that performance on solar panel is higher than the other parts. This is because solar panels are oftentimes clearly separated from the other two parts while antenna and main bodies are 
in many cases embedded with each other.

Because our dataset is the first publicly available space dataset for spacecraft segmentation, it targets different type of objects in a unique scenario when compared to popular segmentation datasets such as Cityscapces~\cite{Cordts2016Cityscapes} or Pascal VOC~\cite{Pascal_VOC}. Nonetheless, for the sake of benchmarking the performance of space-based semantic segmentation against other on-Earth scenarios, we compare performances of state-of-the-art models across different datasets, by treating spacecraft parts as classes of objects for our dataset. 


Table~\ref{tab:parts_as_classes} reports the result of semantic segmentation on 4 datasets: Pascal VOC, Cityscapes, Cityscapes Val and our dataset. 
Overall, the average mIoU of Earth-based datasets are 6\% to 11\% higher compared to that of our dataset. 
It appears that current state-of-the-art models have inferior performance when deployed directly to identify and segment spacecraft parts. 

In Figure~\ref{fig:qua_results} we provide a few qualitative results of our dataset from models in Table~\ref{tab:parts_as_classes}. 
As we can see, the complex structures in spacecrafts can result in precarious predictions of the masks of parts. Additionally, all models struggle to well identify the antenna of the spacecraft in the first row of Figure~\ref{fig:qua_results}, which complies with the mIoU scores in Table~\ref{tab:parts_segmentation}.

Overall, the task of space-based semantic segmentation might not be directly solvable by models designed for on-Earth scenarios. Our dataset thus serves to address this gap by bringing novel challenges in this task.

\section{Conclusion}


We propose a space image dataset for vision-based spacecraft detection and segmentation tasks. Our dataset consists of 3117 space-based images of satellites and space stations, with annotations of object bounding boxes, instance and parts masks. We use a bootstrap strategy during dataset development to reduce manual labors. We conduct experiments in object detection, instance and semantic segmentation using state-of-the-art methods and benchmark our dataset for future space-based vision research.

\bibliographystyle{unsrt}
\bibliography{reference}
\end{multicols}
\end{document}